\documentclass[10pt]{article}
\usepackage{colacl}
\usepackage{epsfig}
\title{Compact non-left-recursive grammars using the 
       selective left-corner transform and factoring\thanks{
       This research was supported by NSF awards 9720368, 9870676 and 9812169.
       We would like to thank our colleagues in BLLIP (Brown Laboratory for
       Linguistic Information Processing) and Bob Moore for their helpful comments on
       this paper.  
  }}
\author{Mark Johnson \qquad Brian Roark \\
        Cognitive and Linguistic Sciences, Box 1978 \\
        Brown University \\
        {\sf Mark\_Johnson@Brown.edu \qquad Brian\_Roark@Brown.edu}}

\newcommand{\XXX}{}

\newcommand{\lc}[2]{#1\raise 0.1 em \hbox{\small --}#2}
\newcommand{\pt}[2]{#1\raise 0.17em \hbox{\tiny$\backslash$}#2}
\newcommand{\LC}{{\cal LC}}
\newcommand{\T}{{\cal T}}

\newcommand{\rewrites}{\;\rightarrow\;}

\newcommand{\LCX}[1]{\large$\displaystyle\mathop{\mbox{\huge$\Rightarrow$}}_{\LC}$}
\newcommand{\EPSREMOVAL}{$\displaystyle\mathop{\mbox{\huge$\Rightarrow$}}_{\mbox{\footnotesize$\epsilon$-removal}}$}

\newcounter{subequation}[equation]
\renewcommand{\thesubequation}{\alph{subequation}}
\newcommand{\twoparteqno}{\mbox{\theequation\thesubequation}}

\makeatletter
\newcommand{\neqno}[1]{ %
  \stepcounter{equation}\stepcounter{subequation}(\twoparteqno) %
  \let\@currentlabel\twoparteqno \label{#1}}

\newcommand{\nneqno}[1]{ %
  \stepcounter{subequation}(\twoparteqno) %
  \let\@currentlabel\twoparteqno \label{#1}}

\newcommand{\nnxform}[3]{#1 & #2 & \nneqno{#3}}
\newcommand{\xform}[3]{#1 & #2}

\long\def\@caption#1[#2]#3{%
  \par
  \addcontentsline{\csname ext@#1\endcsname}{#1}%
    {\protect\numberline{\csname the#1\endcsname}{\ignorespaces #2}}%
  \begingroup
    \@parboxrestore
    \if@minipage
      \@setminipage
    \fi
    \small 
    \@makecaption{\csname fnum@#1\endcsname}{\ignorespaces #3}\par
  \endgroup}

\makeatother

\begin{document}

\maketitle

\begin{abstract}
The left-corner transform removes left-recursion from (probabilistic)
context-free grammars and unification grammars, permitting simple
top-down parsing techniques to be used.  Unfortunately the grammars
produced by the standard left-corner transform are usually much larger
than the original.  The selective left-corner transform described in
this paper produces a transformed grammar which simulates left-corner
recognition of a user-specified set of the original productions, and
top-down recognition of the others.  Combined with two factorizations,
it produces non-left-recursive grammars that are not much larger than
the original.
\end{abstract}

\section{Introduction}
Top-down parsing techniques are attractive because of their
simplicity, and can often achieve good performance in practice
\cite{Roark99b}.  However, with a left-recursive grammar such parsers
typically fail to terminate.  The left-corner grammar transform
converts a left-recursive grammar into a non-left-recursive one: a
top-down parser using a left-corner transformed grammar simulates a
left-corner parser using the original grammar
\cite{Rosenkrantz70,Aho72}.  However, the left-corner transformed
grammar can be significantly larger than the original grammar, causing
numerous problems.  For example, we show below that a probabilistic
context-free grammar (PCFG) estimated from left-corner transformed
Penn WSJ tree-bank trees exhibits considerably greater sparse data
problems than a PCFG estimated in the usual manner, simply because the
left-corner transformed grammar contains approximately 20~times \XXX more
productions.  The transform described in this paper produces a grammar
approximately the same size as the input grammar, which is not as
adversely affected by sparse data. 

Left-corner transforms are particularly useful because they can
preserve annotations on productions (more on this below) and are
therefore applicable to more complex grammar formalisms as well as
CFGs; a property which other approaches to left-recursion elimination
typically lack.  For example, they apply to left-recursive
unification-based grammars \cite{Matsumoto83,Pereira87,Johnson98a}.
Because the emission probability of a PCFG production can be regarded
as an annotation on a CFG production, the left-corner transform can
produce a CFG with weighted productions which assigns the same
probabilities to strings and transformed trees as the original grammar
\cite{Abney99a}.  However, the transformed grammars can be much larger
than the original, which is unacceptable for many applications
involving large grammars.

The selective left-corner transform reduces the transformed grammar
size because only those productions which appear in a left-recursive
cycle need be recognized left-corner in order to remove
left-recursion.  A top-down parser using a grammar produced by the
selective left-corner transform simulates a {\em generalized
left-corner parser} \cite{Demers77,Nijholt80} which recognizes a
user-specified subset of the original productions in a left-corner
fashion, and the other productions top-down.

Although we do not investigate it in this paper, the selective
left-corner transform should usually have a smaller search space
relative to the standard left-corner transform, all else being equal.
The partial parses produced during a top-down parse consist of a
single connected tree fragment, while the partial parses produced
produced during a left-corner parse generally consist of several
disconnected tree fragments.  Since these fragments are only weakly
related (via the ``link'' constraint described below), the search for
each fragment is relatively independent.  This may be responsible for
the observation that exhaustive left-corner parsing is less efficient
than top-down parsing \cite{Covington94}.  Informally, because the
selective left-corner transform recognizes only a subset of
productions in a left-corner fashion, its partial parses contain fewer
tree discontiguous fragments and the search may be more efficient.

While this paper focuses on reducing grammar size to minimize sparse
data problems in PCFG estimation, the modified left-corner transforms
described here are generally applicable wherever the original
left-corner transform is.  For example, the selective left-corner
transform can be used in place of the standard left-corner transform
in the construction of finite-state approximations \cite{Johnson98a},
often reducing the size of the intermediate automata constructed.  The
selective left-corner transform can be generalized to head-corner
parsing \cite{vanNoord97}, yielding a selective head-corner parser.
(This follows from generalizing the selective left-corner
transform to Horn clauses).

After this paper was accepted for publication we learnt of
Moore~\shortcite{Moore00a}, which addresses the issue of grammar size
using very similar techniques to those proposed here.  The goals of
the two papers are slightly different: Moore's approach is designed to
reduce the total grammar size (i.e., the sum of the lengths of the
productions), while our approach minimizes the number of productions.
Moore~\shortcite{Moore00a} does not address left-corner
tree-transforms, or questions of sparse data and parsing accuracy that
are covered in section~\ref{s:results}.


\section{The selective left-corner and related transforms}
This section introduces the selective left-corner transform and two
additional factorization transforms which apply to its output.  These
transforms are used in the experiments described in the following
section.  As Moore~\shortcite{Moore00a} observes, in general the
transforms produce a non-left-recursive output grammar only if the
input grammar $G$ does not contain unary cycles, i.e., there is no
nonterminal $A$ such that $A \rightarrow^{+}_{G} A$. 

\subsection{The selective left-corner transform}
The selective left-corner transform takes as input a CFG $G=
(V,T,P,S)$ and a set of {\em left-corner productions} $L \subseteq P$,
which contains no epsilon productions; the non-left-corner productions
$P-L$ are called {\em top-down productions}.  The {\em standard
left-corner transform} is obtained by setting $L$ to the set of all
non-epsilon productions in $P$.  
The {\em selective left-corner transform} of $G$ with respect
to $L$ is the CFG $\LC_L(G) = (V_1, T, P_1, S)$, where:
\[
 V_1 \; = \; V \cup \{ \lc{D}{X} : D \in V, X \in V \cup T \}
\]
and $P_1$ contains all instances of the schemata \ref{slc1}.  In these
schemata, $D \in V$, $w \in T$, and lower case
greek letters range over $(V \cup T)^*$.  The $\lc{D}{X}$ are new
nonterminals; informally they encode a parse state in which an $D$ is
predicted top-down and an $X$ has been found left-corner, so
   $\lc{D}{X} \mathop{\Rightarrow}_{\LC_L(G)}^* \gamma$
only if $D \mathop{\Rightarrow}_{G}^* X \gamma$.
\[ \refstepcounter{equation}\label{slc1}
\begin{array}{lll}
\nnxform{D \rewrites w \; \lc{D}{w}}{}{slc1a} \\
\nnxform{D \rewrites \alpha \; \lc{D}{A}}{\mbox{where } A \rightarrow \alpha \in P - L}{slc1b} \\
\nnxform{\lc{D}{B} \rewrites \beta \; \lc{D}{C}}{\mbox{where } C \rightarrow B\,\beta \in L}{slc1c} \\
\nnxform{\lc{D}{D} \rewrites \epsilon}{}{slc1d}
\end{array}
\]
The schemata function as follows.  The productions introduced by
schema~\ref{slc1a} start a left-corner parse of a predicted
nonterminal $D$ with its leftmost terminal $w$, while those introduced
by schema~\ref{slc1b} start a left-corner parse of $D$ with a
left-corner $A$, which is itself found by the top-down recognition of
production $A\rightarrow\alpha \in P-L$.  Schema~\ref{slc1c} extends
the current left-corner $B$ up to a $C$ with the left-corner
recognition of production $C\rightarrow B\,\beta$.  Finally,
schema~\ref{slc1d} matches the top-down prediction with the recognized
left-corner category.

\begin{figure}
\begin{center}
\input{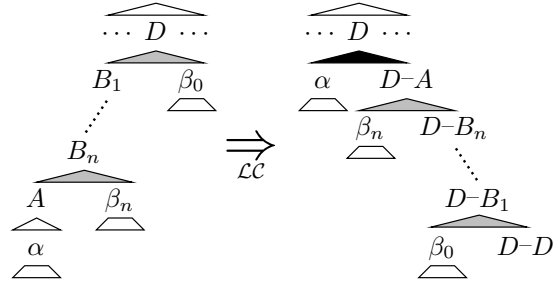}
\vspace*{-1.5em}
\end{center}
\caption{ \label{f:slc}
 Schematic parse trees generated by the original grammar $G$ and the
 selective left-corner transformed grammar $\LC_L(G)$.  The shaded
 local trees in the original parse tree correspond to left-corner
 productions; the corresponding local trees (generated by instances of
 schema~\ref{slc1c}) in the selective left-corner transformed tree
 are also shown shaded.  The local tree colored black is generated by
 an instance of schema~\ref{slc1b}.}
\end{figure}

Figure~\ref{f:slc} schematically depicts the relationship between a
chain of left-corner productions in a parse tree generated by $G$ and
the chain of corresponding instances of schema~\ref{slc1c}.  The
left-corner recognition of the chain starts with the recognition of
$\alpha$, the right-hand side of a top-down production
$A\rightarrow\alpha$, using an instance of schema~\ref{slc1b}.  The
left-branching chain of left-corner productions corresponds to a
right-branching chain of instances of schema~\ref{slc1c}; the left-corner
transform in effect converts left recursion into right recursion.
Notice that the top-down predicted category $D$ is passed down
this right-recursive chain, effectively multiplying each left-corner
productions by the possible top-down predicted categories.
The right recursion terminates with an instance of schema~\ref{slc1d}
when the left-corner and top-down categories match. 

\begin{figure}
\begin{center}
\input{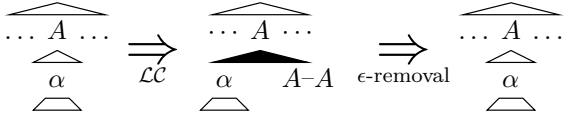}
\vspace*{-1.5em}
\end{center}
\caption{ \label{f:slctd}
 The recognition of a top-down production $A\rightarrow\alpha$ 
 by $\LC_L(G)$ involves a left-corner category $\lc{A}{A}$,
 which immediately rewrites to $\epsilon$.  One-step $\epsilon$-removal applied to
 $\LC_L(G)$ produces a grammar in which each top-down production
 $A\rightarrow\alpha$ corresponds to a production $A\rightarrow\alpha$
 in the transformed grammar.}
\end{figure}

Figure~\ref{f:slctd} shows how top-down productions from $G$ are recognized
using $\LC_L(G)$.  When the selective left-corner transform is
followed by a one-step $\epsilon$-removal transform (i.e., composition
or partial evaluation of schema~\ref{slc1b} with respect to
schema~\ref{slc1d} \cite{Johnson98a,Abney91,Resnik92}), each top-down
production from $G$ appears unchanged in the final grammar. 
Full $\epsilon$-removal yields the grammar given by the schemata below.
\[ \refstepcounter{equation}\label{slc1eps}
\begin{array}{lll}
\xform{D \rewrites w \; \lc{D}{w}}{}{slc1epsa1} \\
\xform{D \rewrites w}{\mbox{where } D \Rightarrow_L^{+} w}{slc1epsa2} \\
\xform{D \rewrites \alpha \; \lc{D}{A}}{\mbox{where } A \rightarrow \alpha \in P - L}{slc1epsb1} \\
\xform{D \rewrites \alpha}{\mbox{where } D \Rightarrow_L^{\star} A, A \rightarrow \alpha \in P - L}{slc1epsb2} \\
\xform{\lc{D}{B} \rewrites \beta \; \lc{D}{C}}{\mbox{where } C \rightarrow B\,\beta \in L}{slc1c1} \\
\xform{\lc{D}{B} \rewrites \beta}{\mbox{where } D \Rightarrow_L^{\star} C, C \rightarrow B\,\beta \in L}{slc1c2}
\end{array}
\]

Moore~\shortcite{Moore00a} introduces a version of the left-corner transform
called ${\rm LC}_{\it LR}$, which applies only to productions with
left-recursive parent and left child categories.  In the context of the other
transforms that Moore introduces, it seems to have the same effect in his system
as the selective left-corner transform does here.

\subsection{Selective left-corner tree transforms}
There is a 1-to-1 correspondence between the parse trees generated by
$G$ and $\LC_L(G)$.  A tree $t$ is generated by $G$ iff there is a
corresponding $t'$ generated by $\LC_L(G)$, where each occurrence of a
top-down production in the derivation of $t$ corresponds to exactly
one local tree generated by occurrence of the corresponding instance of
schema~\ref{slc1b} in the derivation of $t'$, and each occurrence of a
left-corner production in $t$ corresponds to exactly one occurrence of
the corresponding instance of schema~\ref{slc1c} in $t'$.  It is
straightforward to define a 1-to-1 tree transform $\T_L$ mapping parse
trees of $G$ into parse trees of $\LC_L(G)$
\cite{Johnson98a,Roark99b}.  In the empirical evaluation below, we
estimate a PCFG from the trees obtained by applying $\T_L$ to the
trees in the Penn WSJ tree-bank, and compare it to the PCFG estimated
from the original tree-bank trees.  A stochastic top-down parser using
the PCFG estimated from the trees produced by $\T_L$ simulates a
stochastic generalized left-corner parser, which is a generalization
of a standard stochastic left-corner parser that permits productions
to be recognized top-down as well as left-corner \cite{Manning97}.
Thus investigating the properties of PCFG estimated from trees transformed
with $\T_L$ is an easy way of studying stochastic push-down automata performing
generalized left-corner parses.

\subsection{Pruning useless productions} \label{s:linkage}
We turn now to the problem of reducing the size of the grammars produced
by left-corner transforms.
Many of the productions generated by schemata~\ref{slc1} are useless, i.e.,
they never appear in any terminating derivation.  While they can be
removed by standard methods for deleting useless productions
\cite{Hopcroft79}, the relationship between the parse
trees of $G$ and $\LC_L(G)$ depicted in Figure~\ref{f:slc} shows how 
to determine ahead of time the new nonterminals
$\lc{D}{X}$ that can appear in useful productions of $\LC_L(G)$.
This is known as a {\em link constraint}.

For (P)CFGs there is a particularly simple link constraint:
$\lc{D}{X}$ appears in useful productions of $\LC_L(G)$ only if
$\exists \gamma \in (V\cup T)^{\star} .\, D \Rightarrow_L^{\star} X \gamma$.
If epsilon removal is applied to the resulting grammar, 
$\lc{D}{X}$ appears in useful productions only if
$\exists \gamma \in (V\cup T)^{+} .\, D \Rightarrow_L^{\star} X \gamma$.
Thus one only need generate instances of the left-corner schemata
which satisfy the corresponding link constraints.

Moore~\shortcite{Moore00a} suggests an additional constraint on
nonterminals $\lc{D}{X}$ that can appear in useful productions of
$\LC_L(G)$: $D$ must either be the start symbol of $G$ or else appear
in a production $A \rightarrow \alpha D \beta$ of $G$, for any $A \in
V$, $\alpha \in \{ V \cup T \}^{+}$ and $\beta \in \{ V \cup T
\}^{\star}$.  It is easy to see that the productions that Moore's
constraint prohibits are useless.  There is one non-terminal in the
tree-bank grammar investigated below that has this property, namely
{\sf\small LST}.  However, in the tree-bank grammar none of the
productions expanding {\sf\small LST} are left-recursive (in fact, the
first child is always a preterminal), so Moore's constraint does not
affect the size of the transformed grammars investigated below.

While these constraints can dramatically reduce both the number of
productions and the size of the parsing search space of the
transformed grammar, in general the transformed grammar $\LC_L(G)$
can be quadratically larger than $G$.  There are two causes for the
explosion in grammar size.  First, $\LC_L(G)$ contains an instance
of schema~\ref{slc1b} for each top-down production
$A\rightarrow\alpha$ and each $D$ such that $\exists \gamma .\, D
\Rightarrow_{L}^{\star} A \gamma$.  Second, $\LC_L(G)$ contains an instance
of schema~\ref{slc1c} for each left-corner production
$C\rightarrow\beta$ and each $D$ such that $\exists \gamma .\, D\Rightarrow_{L}^{\star}
C \gamma$.  In effect, $\LC_L(G)$ contains one copy of each production for
each possible left-corner ancestor.  Section~\ref{s:factor} describes
further factorizations of the productions of $\LC_L(G)$ which
mitigate these causes.

\subsection{Optimal choice of $L$} \label{s:optL}
Because $\Rightarrow_{L}^{\star}$ increases monotonically with
$\Rightarrow_{L}$ and hence $L$, we typically reduce the size of
$\LC_L(G)$ by making the left-corner production set 
$L$ as small as possible.  This section shows
how to find the unique minimal set of left-corner productions $L$
such that $\LC_L(G)$ is not left-recursive.

Assume $G=(V,T,P,S)$ is pruned (i.e., $P$ contains no useless
productions) and that there is no $A\in V$ such that
$A\rightarrow_{G}^{+} A$ (i.e., $G$ does not generate recursive unary
branching chains).  For reasons of space we also assume that $P$
contains no $\epsilon$-productions, but this approach can be extended
to deal with them if desired.  A production $A\rightarrow B\beta\in P$
is {\em left-recursive} iff $\exists \gamma \in (V\cup T)^{\star} .\,
B\Rightarrow_{P}^{\star} A \gamma$, i.e., $P$ rewrites $B$ into a string
beginning with $A$.  Let $L_0$ be the set of left-recursive
productions in $G$.  Then we claim (1) that $\LC_{L_0}(G)$ is not
left-recursive, and (2) that for all $L\subset L_0$, $\LC_L(G)$ is
left-recursive.

Claim~1 follows from the fact that if $A \Rightarrow_{L_0} B \gamma$
then $A \Rightarrow_{P} B \gamma$
and the constraints in section~\ref{s:linkage} on useful
productions of $\LC_{L_0}(G)$.  Claim~2 follows from the fact that
if $L\subset L_0$ then there is a chain of left-recursive productions
that includes a top-down production; a simple induction on the length
of the chain shows that $\LC_L(G)$ is left-recursive.

This result justifies the common practice in natural language
left-corner parsing of taking the terminals to be the preterminal
part-of-speech tags, rather than the lexical items themselves.  (We
did not attempt to calculate the size of such a left-corner grammar in
the empirical evaluation below, but it would be much larger than any
of the grammars described there).  In fact, if the preterminals are
distinct from the other nonterminals (as they are in the tree-bank
grammars investigated below) then $L_0$ does not include any
productions beginning with a preterminal, and $\LC_{L_0}(G)$ contains
no instances of schema~\ref{slc1a} at all.  We now turn our attention
to the other schemata of the selective left-corner grammar transform.

\subsection{Factoring the output of $\LC_L$} \label{s:factor}
This section defines two factorizations of the output of the selective
left-corner grammar transform that can dramatically reduce its size.
These factorizations are most effective if the number of productions
is much larger than the number of nonterminals, as is usually the case with
tree-bank grammars.

The {\em top-down factorization} decomposes schema~\ref{slc1b} by
introducing new nonterminals $D'$, where $D \in V$, that have the same
expansions that $D$ does in $G$.  Using the same interpretation for
variables as in schemata~\ref{slc1}, if $G=(V,T,P,S)$ then
$\LC_L^{(td)}(G) = (V_{td},T,P_{td},S)$, where:
\[
 V_{td} \; = \; V_1 \cup \{ D' : D \in V \}
\]
and $P_{td}$ contains all instances of the schemata~\ref{slc1a},
\ref{slc2a}, \ref{slc2b}, \ref{slc1c} and~\ref{slc1d}.
\[ \refstepcounter{equation}\label{slc2}
\begin{array}{lll}
\nnxform{D \rewrites A' \; \lc{D}{A}}{\mbox{where } A \rightarrow \alpha \in P - L}{slc2a} \\
\nnxform{A' \rewrites \alpha}{\mbox{where } A \rightarrow \alpha \in P-L}{slc2b} 
\end{array}
\]
Notice that the number of instances of schema~\ref{slc2a} is less than
the square of the number of nonterminals and that the number of
instances of schema~\ref{slc2b} is the number of top-down productions;
the sum of these numbers is usually much less than the number of
instances of schema~\ref{slc1b}.

Top-down factoring plays approximately the same role as
``non-left-recursion grouping'' (NLRG) does in
Moore's~\shortcite{Moore00a} approach.  The major difference is that
NLRG applies to all productions $A \rightarrow B \beta$ in which $B$
is not left-recursive, i.e., $\not\!\exists\gamma .\, B \Rightarrow_{P}^{+} B \gamma$, while
in our system top-down factorization applies to those productions
for which $\not\!\exists\gamma .\, B \Rightarrow_{P}^{\star} A \gamma$,
i.e., the productions not directly involved in left recursion.

The {\em left-corner factorization} decomposes schema~\ref{slc1c}
in a similar way using new nonterminals $\pt{D}{X}$, where
$D \in V$ and $X \in V \cup T$.  $\LC_L^{(lc)}(G) = (V_{lc},T,P_{lc},S)$,
where:
\[
 V_{lc} \; = \; V_1 \cup \{ \pt{D}{X} : D \in V, X \in V \cup T \}
\]
and $P_{lc}$ contains all instances of the schemata~\ref{slc1a},
\ref{slc1b}, \ref{slc3a}, \ref{slc3b} and~\ref{slc1d}.
\[ \refstepcounter{equation}\label{slc3}
\begin{array}{lll}
\nnxform{\lc{D}{B} \rewrites \pt{C}{B} \; \lc{D}{C}}{\mbox{where } C \rightarrow B\,\beta \in L}{slc3a} \\
\nnxform{\pt{C}{B} \rewrites \beta}{\mbox{where } C \rightarrow B\,\beta \in L}{slc3b} 
\end{array}
\]
The number of instances of schema~\ref{slc3a} is bounded by the number
of instances of schema~\ref{slc1c} and is typically much smaller,
while the number of instances of schema~\ref{slc3b} is precisely the
number of left-corner productions $L$.

Left-corner factoring seems to correspond to one step of
Moore's~\shortcite{Moore00a} ``left factor'' (LF) operation.  The left
factor operation constructs new nonterminals corresponding to common
prefixes of arbitrary length, while left-corner factoring effectively
only factors the first nonterminal symbol on the right hand side of
left-corner productions.  While we have not done experiments, Moore's
left factor operation would seem to reduce the total number of symbols
in the transformed grammar at the expense of possibly introducing
additional productions, while our left-corner factoring reduces the
number of productions.

These two factorizations can be used together in the obvious way to
define a grammar transform $\LC_L^{(td,lc)}$, whose productions are
defined by schemata~\ref{slc1a}, \ref{slc2a}, \ref{slc2b},
\ref{slc3a}, \ref{slc3b} and~\ref{slc1d}.  There are corresponding
tree transforms, which we refer to as $\T_L^{(td)}$, etc., below.  Of
course, the pruning constraints described in section~\ref{s:linkage}
are applicable with these factorizations, and corresponding invertible
tree transforms can be constructed.


\section{Empirical Results} \label{s:results}
To examine the effect of the transforms outlined above, we
experimented with various PCFGs induced from sections~2--21 of a
modified Penn WSJ tree-bank as described in
Johnson~\shortcite{Johnson98c} (i.e., labels simplified to grammatical
categories, {\sc root} nodes added, empty nodes and vacuous unary
branches deleted, and auxiliaries retagged as {\sc aux} or {\sc
auxg}).  We ignored lexical items, and treated the part-of-speech tags
as terminals.  As Bob Moore pointed out to us, the left-corner
transform may produce left-recursive grammars if its input grammar
contains unary cycles, so we removed them using the a transform that Moore
suggested.  Given an initial set of (non-epsilon) productions $P$,
the transformed grammar contains the following productions, where
the $A^\natural$ are new non-terminals:
\[\begin{array}{ll}
A \rightarrow \alpha & \mbox{where } A \rightarrow \alpha \in P, A \not\Rightarrow_{P}^{+} A \\
A \rightarrow D^\natural & \mbox{where } A \Rightarrow_{P}^{\star} D \Rightarrow_{P}^{+} A \\
A^\natural \rightarrow \alpha & \mbox{where } A \rightarrow \alpha \in P, A \Rightarrow_P^{+} A, \alpha \not\Rightarrow_{P}^{*} A
\end{array}\]
This transform can be extended to one on PCFGs which preserves
derivation probabilities.  In this section, we fix $P$ to be the
productions that result after applying this unary cycle removal
transformation to the tree-bank productions, and $G$ to be
the corresponding grammar.

Tables~\ref{t:1} and~\ref{t:1e} give the sizes of selective
left-corner grammar transforms of $G$ for various values of the
left-corner set $L$ and factorizations, without and with
epsilon-removal respectively.  In the tables, $L_0$ is the set of
left-recursive productions in $P$, as defined in section~\ref{s:optL}.
$N$ is the set of productions in $P$ whose left-hand sides do not
begin with a part-of-speech (POS) tag; because POS tags are distinct
from other nonterminals in the tree-bank, $N$ is an easily identified
set of productions guaranteed to include $L_0$.  The tables also gives
the sizes of maximum-likelihood PCFGs estimated from the trees
resulting from applying the selective left-corner tree transforms $\T$
to the tree-bank, breaking unary cycles as described above.  
For the parsing experiments below we always deleted
empty nodes in the output of these tree transforms; this corresponds
to epsilon removal in the grammar transform.  

\begin{table}
\begin{tabular}{c|rrrr}
 & \multicolumn{1}{|c}{none} & \multicolumn{1}{c}{$(td)$} &
 \multicolumn{1}{c}{$(lc)$} & \multicolumn{1}{c}{$(td,lc)$} \\ \hline
 $G$ & 15,040 & & & \vbox to 1.1em {} \\ 
 $\LC_P$ & 346,344 & & 30,716 & \vbox to 1.1em {} \\ 
 $\LC_N$ & 345,272 & 113,616 & 254,067 & 22,411 \\ 
 $\LC_{L_0}$ & 314,555 & 103,504 & 232,415 & 21,364 \\
 $\T_P$ & 20,087 & & 17,146 & \vbox to 1.1em {} \\ 
 $\T_N$ & 19,619 & 16,349 & 19,002 & 15,732 \\ 
 $\T_{L_0}$ & 18,945 & 16,126 & 18,437 & 15,618
\end{tabular}
\caption{ \label{t:1}
 Sizes of PCFGs inferred using various grammar and tree transforms
 after pruning with link constraints without epsilon removal.
 Columns indicate factorization.  In the grammar and tree transforms,
 $P$ is the set of productions in $G$ (i.e., the standard
 left-corner transform), $N$ is the set of all productions in $P$ which
 do not begin with a POS tag, and $L_0$ is the set of
 left-recursive productions.}
\end{table}

\begin{table}
\begin{tabular}{c|rrrr}
 & \multicolumn{1}{|c}{none} & \multicolumn{1}{c}{$(td)$} &
 \multicolumn{1}{c}{$(lc)$} & \multicolumn{1}{c}{$(td,lc)$} \\ \hline
 $\LC_P$ & 564,430 & & 38,489 & \vbox to 1.1em {} \\ 
 $\LC_N$ & 563,295 & 176,644 & 411,986 & 25,335 \\ 
 $\LC_{L_0}$ & 505,435 & 157,899 & 371,102 & 23,566 \\
 $\T_P$ & 22,035 & & 17,398 & \vbox to 1.1em {} \\ 
 $\T_N$ & 21,589 & 16,688 & 20,696 & 15,795 \\ 
 $\T_{L_0}$ & 21,061 & 16,566 & 20,168 & 15,673
\end{tabular}
\caption{ \label{t:1e}
 Sizes of PCFGs inferred using various grammar and tree transforms
 after pruning with link constraints with epsilon removal, using
 the same notation as Table~\ref{t:1}.}
\end{table}

First, note that $\LC_P(G)$, the result of applying the standard
left-corner grammar transform to $G$, has approximately 20~times \XXX the
number of productions that $G$ has.  However $\LC_{L_0}^{(td,lc)}(G)$,
the result of applying the selective left-corner grammar
transformation with factorization, has approximately 1.4~times \XXX the
number of productions that $G$ has.  Thus the methods described in
this paper can in fact dramatically reduce the size of left-corner
transformed grammars.  Second, note that $\LC_N^{(td,lc)}(G)$ is not
much larger than $\LC_{L_0}^{(td,lc)}(G)$.  This is because $N$ is not
much larger than $L_0$, which in turn is because most pairs of non-POS
nonterminals $A,B$ are mutually left-recursive.

Turning now to the PCFGs estimated after applying tree transforms, we
notice that grammar size does not increase nearly so dramatically.
These PCFGs encode a maximum-likelihood estimate of the state
transition probabilities for various stochastic generalized
left-corner parsers, since a top-down parser using these grammars
simulates a generalized left-corner parser.  The fact that $\LC_P(G)$
is 17~times \XXX larger than the PCFG inferred after applying $\T_P$ to the
tree-bank means that most of the possible transitions of a standard
stochastic left-corner parser are not observed in the tree-bank
training data.  The state of a left-corner parser does capture some
linguistic generalizations \cite{Manning97,Roark99b}, but one might
still expect sparse-data problems.  Note that $\LC_{L_0}^{(td,lc)}$ is
only 1.4~times \XXX larger than $\T_{L_0}^{(td,lc)}$, so we expect
less serious sparse data problems with the factored selective
left-corner transform.

We quantify these sparse data problems in two ways using a held-out
test corpus, viz., all sentences in section~23 of
the tree-bank.  First, table~\ref{t:noparse} lists the number of
sentences in the test corpus that fail to receive a parse with the
various PCFGs mentioned above.  This is a relatively crude measure,
but correlates roughly with the ratios of grammar sizes, as expected.

Second, table~\ref{t:missrule} lists the number of productions found
in the tree-transformed test corpus that do not appear in the
correspondingly transformed trees of sections~2--21.  What is striking
here is that the number of missing productions after either of the
transforms $\T_{L_0}^{(td,lc)}$ or $\T_N^{(td,lc)}$ is approximately
the same as the number of missing productions using the untransformed
trees, indicating that the factored selective left-corner transforms
cause little or no additional sparse data problem.  (The relationship
between local trees in the parse trees of $G$ and $\LC_L(G)$ mentioned
earlier implies that left-corner tree transformations will not
decrease the number of missing productions).

\begin{table}
\begin{center}
\begin{tabular}{c|cccc}
Transform & \multicolumn{1}{|c}{none} & \multicolumn{1}{c}{$(td)$} &
 \multicolumn{1}{c}{$(lc)$} & \multicolumn{1}{c}{$(td,lc)$} \\ \hline
none & 0 & & &  \\
$\T_{P}$ & 2 & & 0 &  \\
$\T_{N}$ & 2 & 0 & 2 & 0 \\
$\T_{L_0}$ & 0 & 0 & 0 & 0
\end{tabular}
\end{center}
\vspace*{-1em}
\caption{ \label{t:noparse}
 The number of sentences in section~23 that do not
 receive a parse using various grammars estimated from sections~2--21.}
\end{table}

\begin{table}
\begin{center}
\begin{tabular}{c|cccc}
Transform & \multicolumn{1}{|c}{none} & \multicolumn{1}{c}{$(td)$} &
 \multicolumn{1}{c}{$(lc)$} & \multicolumn{1}{c}{$(td,lc)$} \\ \hline
none & 514 & & &  \\
$\T_{P}$ & 665 & & 535 & \\
$\T_{N}$ & 664 & 543 & 639 & 518 \\
$\T_{L_0}$ & 640 & 547 & 615 & 522 \\
$\T_{P\,\epsilon}$ & 719 & & 539 & \\
$\T_{N\,\epsilon}$ & 718 & 554 & 685 & 521 \\
$\T_{L_0\,\epsilon}$ & 706 & 561 & 666 & 521 \\
\end{tabular}
\end{center}
\vspace*{-1em}
\caption{ \label{t:missrule}
 The number of productions found in the transformed trees of sentences
 in section~23 that do not appear in the corresponding
 transformed trees from sections~2--21.  (The subscript epsilon indicates
 epsilon removal was applied).}
\end{table}

We also investigate the accuracy of the maximum-likelihood parses
(MLPs) obtained using the PCFGs estimated from the output of the
various left-corner tree transforms.\footnote{
We did not investigate
the grammars produced by the various left-corner grammar transforms.
Because a left-corner grammar transform $\LC_L$
preserves production probabilities, the highest scoring parses obtained
using the weighted CFG $\LC_L(G)$ should be the highest scoring parses
obtained using $G$ transformed by $\T_L$.}
We searched for these parses
using an exhaustive CKY parser.  Because the parse trees of these
PCFGs are isomorphic to the derivations of the corresponding
stochastic generalized left-corner parsers, we are in fact evaluating
different kinds of stochastic generalized left-corner parsers inferred
from sections~2--21 of the tree-bank.  We used the
transform-detransform framework described in
Johnson~\shortcite{Johnson98c} to evaluate the parses, i.e., we
applied the appropriate inverse tree transform $\T^{-1}$ to
detransform the parse trees produced using the PCFG estimated from
trees transformed by $\T$.  By calculating the labelled precision and
recall scores for the detransformed trees in the usual manner, we can
systematically compare the parsing accuracy of different kinds of
stochastic generalized left-corner parsers.

\begin{table}
\begin{tabular}{c|cccc}
  & \multicolumn{1}{|c}{none} & \multicolumn{1}{c}{$(td)$} &
 \multicolumn{1}{c}{$(lc)$} & \multicolumn{1}{c}{$(td,lc)$} \\ \hline
none & \footnotesize 70.8,75.3 & & &  \\
$\T_{P}$ & \footnotesize 75.8,77.7 & & \footnotesize 74.8,76.9 &  \\
$\T_{N}$ & \footnotesize 75.8,77.6 & \footnotesize 73.8,75.8 & \footnotesize 75.5,77.8 & \footnotesize 72.8,75.4\\
$\T_{L_0}$ & \footnotesize 75.8,77.4 & \footnotesize 73.0,74.7 & \footnotesize75.6,77.8 & \footnotesize 72.9,75.4
\end{tabular}
\caption{ \label{t:precrec}
 Labelled recall and precision scores of PCFGs estimated using various
 tree-transforms in a transform-detransform framework using test
 data from section~23.}
\end{table}

Table~\ref{t:precrec} presents the results of this comparison.  As
reported previously, the standard left-corner grammar embeds
sufficient non-local information in its productions to significantly
improve the labelled precision and recall of its MLPs with respect to
MLPs of the PCFG estimated from the untransformed trees
\cite{Manning97,Roark99b}.  Parsing accuracy drops off as grammar
size decreases, presumably because smaller PCFGs have fewer adjustable
parameters with which to describe this non-local information.
There are other kinds of non-local information which can be incorporated
into a PCFG using a transform-detransform approach that result in an even
greater improvement of parsing accuracy~\cite{Johnson98c}.
Ultimately, however, it seems that a more complex approach incorporating
back-off and smoothing is necessary in order to achieve the parsing accuracy
achieved by Charniak~\shortcite{Charniak97} and Collins~\shortcite{Collins97}.

\section{Conclusion}
This paper presented factored selective left-corner grammar
transforms.  These transforms preserve the primary benefits of the
left-corner grammar transform (i.e., elimination of left-recursion and
preservation of annotations on productions) while dramatically
ameliorating its principal problems (grammar size and sparse data
problems).  This should extend the applicability of left-corner
techniques to situations involving large grammars.  We showed how to
identify the minimal set $L_0$ of productions of a grammar that must
be recognized left-corner in order for the transformed grammar not to
be left-recursive.  We also proposed two factorizations of the output
of the selective left-corner grammar transform which further reduce
grammar size, and showed that there is only a minor increase in
grammar size when the factored selective left-corner transform is
applied to a large tree-bank grammar.  Finally, we exploited the
tree transforms that correspond to these grammar transforms to
formulate and study a class of stochastic generalized left-corner
parsers.

This work could be extended in a number of ways.  For example,
in this paper we assumed that one would always choose a left-corner
production set that includes the minimal set $L_0$ required
to ensure that the transformed grammar is not left-recursive.
However, Roark and Johnson~\shortcite{Roark99b} report good performance
from a stochastically-guided top-down parser, suggesting that left-recursion
is not always fatal.  It might be
possible to judiciously choose a left-corner production set {\em smaller}
than $L_0$ which eliminates pernicious left-recursion, so that the
remaining left-recursive cycles have such low probability that they will
effectively never be used and a stochastically-guided top-down parser
will never search them.

\bibliographystyle{acl}
\small
\bibliography{mj}

\end{document}